\documentclass[twoside,leqno,twocolumn]{article}

\usepackage[letterpaper]{geometry}

\usepackage{ltexpprt}
\usepackage{hyperref}
\usepackage{graphicx}
\usepackage{cite}
\usepackage{amsmath,amssymb,amsfonts}
\usepackage{algorithmic}
\usepackage{graphicx}
\usepackage{textcomp}
\usepackage{xcolor}
\usepackage{booktabs}
\usepackage{bbm}
\usepackage{multirow}
\newcommand{\red}[1]{\textcolor{red}{#1}}

\begin{document}

%


\title{\Large A Linkage-based Doubly Imbalanced Graph Learning Framework \\ for Face Clustering}

\author{Huafeng Yang \thanks{Alibaba Group, Hangzhou, China.} \textsuperscript{~}\thanks{fduspectre@gmail.com, qjshenxdu@gmail.com}
\and Qijie Shen  \footnotemark[1] \footnotemark[2]
\and Xingjian Chen \footnotemark[1] 
\and Fangyi Zhang 
\thanks{Queensland University of Technology, Brisbane, Australia.} \textsuperscript{~}\thanks{fangyi.zhang@qut.edu.au}
\and Rong Du \footnotemark[1]
}


\date{}

\maketitle


\fancyfoot[R]{\scriptsize{Copyright \textcopyright\ 2023 by SIAM\\
Unauthorized reproduction of this article is prohibited}}





\begin{abstract} \small\baselineskip=9pt 
In recent years, benefiting from the expressive power of Graph Convolutional Networks (GCNs), significant breakthroughs have been made in face clustering area. However, rare attention has been paid to GCN-based clustering on imbalanced data. Although imbalance problem has been extensively studied, the impact of imbalanced data on GCN-based linkage prediction task is quite different, which would cause problems in two aspects: imbalanced linkage labels and biased graph representations. 
The former is 
similar to that in classic image classification task, but the latter is a particular problem in GCN-based clustering via linkage prediction. Significantly biased graph representations in training can cause catastrophic over-fitting of a GCN model. 
To tackle these challenges, we propose a linkage-based doubly imbalanced graph learning framework for face clustering.
In this framework, we evaluate the feasibility of those existing methods for imbalanced image classification problem on GCNs, and present a new method to alleviate the imbalanced labels and also augment graph representations using a Reverse-Imbalance Weighted Sampling (RIWS) strategy. With the RIWS strategy, probability-based class balancing weights could ensure the overall distribution of positive and negative samples; In addition, weighted random sampling provides diverse subgraph structures, which effectively alleviates the over-fitting problem
and improves the representation ability of GCNs. Extensive experiments on series of imbalanced benchmark datasets synthesized from MS-Celeb-1M and DeepFashion demonstrate the effectiveness and generality of our proposed method. 
Our implementation and the synthesized datasets will be openly available on
\red{\url{https://github.com/espectre/GCNs_on_imbalanced_datasets}}.
\end{abstract}
\leavevmode

\section{Introduction.}
Face clustering is widely used in many applications such as face retrieval, face annotation and album classification. It aims to group together face images from a certain person in an unsupervised manner. Traditional clustering methods normally assume oversimplified data distribution~\cite{wang2019linkage} that differ a lot from the distribution of large scale face images in the real world, therefore can hardly obtain satisfying performance. 

In recent years, benefiting from the expressive power of Graph Convolutional Networks (GCNs), better performance has been obtained in large scale face clustering benchmarks like MS-Celeb-1M~\cite{guo2016ms} by GCN-based solutions where GCNs
are used for graph, node or edge recognition tasks and also feature embedding.
L-GCN~\cite{wang2019linkage} uses a GCN to predict whether a link exists between a ``pivot'' node and its 1-hop neighbors.
Two GCNs are used in \cite{yang2019learninggcnds} to detect and segment
cluster proposals. \cite{yang2020learninggcnve} also uses two GCNs to
complete face clustering: one to estimate the confidence of vertices, and
the other to measure the connectivity across vertices.
DA-Net~\cite{guo2020density} leverages both local and non-local information
to obtain better feature embedding. 

\begin{figure}[!t]
  \centering
  \includegraphics[width=0.4\textwidth]{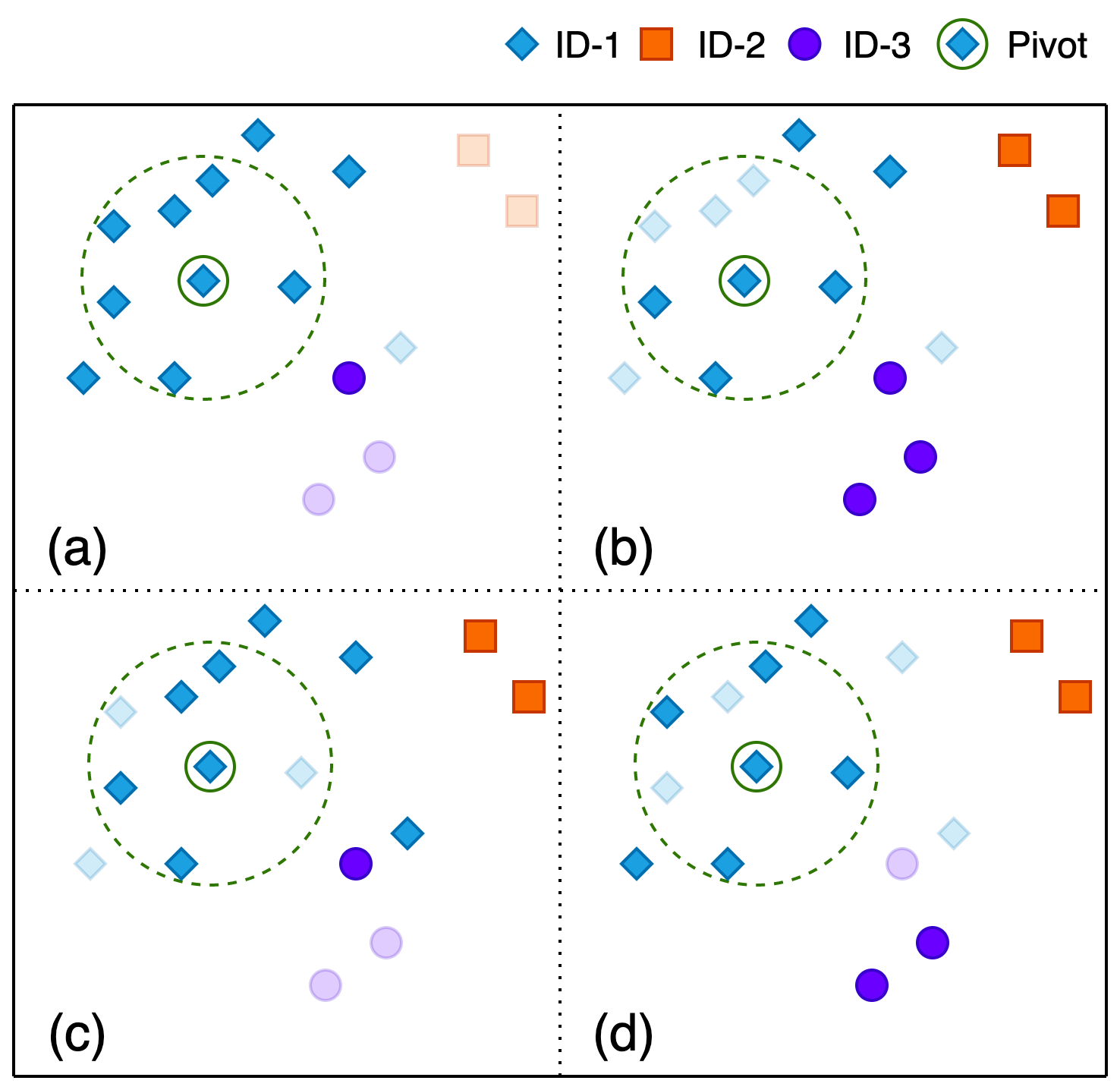}
  \caption{1-hop nodes of subgraphs contructed by RIWS. In each sub-picture, all nodes represent the candidate 1-hop nodes obtained by the pivot's eKNN, and the nodes with high transparency means that they are selected to construct a subgraph.}
  \label{Fig.1}
\end{figure}

Although the aforementioned methods have made significant progress, their performance is often greatly compromised when facing imbalanced data in real scenes (i.e., numbers of face samples for different persons vary in a large scale with an imbalanced distribution). 
To help reduce this kind of performance compromise, this paper particularly studies imbalance problems in GCN-based linkage prediction task (taking the L-GCN method as an exemplary baseline), where rare attention has been paid. 

In the past years, imbalance problems in image classification, whose nature lies in imbalanced numbers of positive and negative samples (i.e., imbalance in labels), have been extensively studied~\cite{cui2019class, kang2019decoupling, zhou2020bbn}.
In L-GCN, the same nature exists in the form of 
imbalanced positive and negative links. However, 
apart from the imbalance problem in labels, 
imbalanced data can also cause biased graph 
representations in L-GCN, which is particularly 
related to GCNs. 
Specifically, graphs generated from an imbalanced training set are prone to having imbalanced structures (with imbalanced numbers of same-class and different-class nodes and also edges),
which is not good for model generalization (i.e., obtaining models that can handle any graph structure). 

Therefore, we propose a linkage-based Doubly Imbalanced Graph Learning (DIGL) framework for face clustering, which divides the clustering process into two stages: subgraph generation and cluster generation.
Some typical approaches to the imbalance problem for image classification are firstly adopted to address the imbalance problem in labels, with some of which show their effectiveness. To address the problems of both imbalanced labels and biased graph representations, a Reverse-Imbalance Weighted Sampling (RIWS) strategy is proposed in this paper to augment graph representations by providing more diverse structures yet maintaining a balanced overall distribution on training samples.

Fig.~\ref{Fig.1} shows some of the typical subgraph structures (mainly 1-hop nodes with their edges ignored for easier reading) constructed by RIWS, where both balanced and imbalanced structures are covered. In comparison, subgraphs constructed by the original L-GCN tend to be extremely imbalanced (Fig.~\ref{Fig.1} (a)), while subgraphs generated via normal re-sampling approaches are all prone to having absolutely balanced structures (Fig.~\ref{Fig.1} (b)), both of which are biased, towards either imbalanced or balanced.
The effectiveness of the RIWS strategy is demonstrated in both face clustering (MS-Celeb-1M) and clothes clustering (DeepFashion), where steady performance gains are obtained.

In summary, this paper has four major contributions:
(1) To the best of our knowledge, two levels of imbalance problems (imbalanced linkage labels and biased graph representations) in GCN-based linkage prediction tasks are studied for the first time. (2) A linkage-based Doubly Imbalanced Graph Learning (DIGL) framework for face clustering is proposed. (3) Typical re-sampling and re-weighting approaches for the imbalance problem in image classification are transferred to tackle the label imbalance problem in GCN-based linkage prediction task, with evaluations on their effectiveness and insightful analyses. (4) A novel strategy named RIWS is proposed to tackle the problems of both imbalanced labels and biased graph representations by increasing the diversity of graph structures yet maintaining a balanced overall distribution on training samples.

\section{Related Work}

\subsection{GCN-based face clustering}
\leavevmode
\newline
Face clustering is essential for exploiting unlabeled face data, and has been widely used in many scenarios. Traditional methods, such as K-Means ~\cite{lloyd1982leastkmeans}, DBSCAN ~\cite{ester1996densitydbscan} and HAC~\cite{sibson1973slinkhac}, are first applied in face clustering task. However, due to some naive assumptions (e.g, same density or convex shape for all clusters), these methods can not handle large scale face data in real world~\cite{wang2019linkage}. In recent years, Graph Convolution Networks (GCNs) is becoming an increasingly powerful technique for clustering, and has achieved significant performance improvement. The graph nature of GCN makes it superior to solve non-Euclidean data related tasks. Recently, considerable research effort has been devoted to solving face clustering with GCNs,  since it can capture the complex relationship between different faces.

L-GCN~\cite{wang2019linkage} formulates face clustering as a linkage prediction problem. If two faces are predicted to be linked, they are clustered together. In ~\cite{yang2019learninggcnds}, two GCN modules, namely GCN-D (detection) and GCN-S (segmentation), are exploited to cluster faces. It is a two-stage procedure, where GCN-D is utilized to select high-quality cluster proposals, and GCN-S is used to remove noises in the proposals. 
~\cite{yang2020learninggcnve} is also a two-stage solution. 
GCN-V (vertex) estimates the confidence of all vertices, and only vertices with higher confidence are selected to construct subgraph. GCN-E (edge) serves as a connectivity estimator.
Ada-NETS ~\cite{wang2022ada} proposes an adaptive neighbor discovery strategy to determine the number of edges connected to each face image.
DA-Net ~\cite{guo2020density} exploits local and global information through clique and chain to obtain better feature embedding.
\subsection{Imbalanced learning}
\leavevmode
\newline
There are already numerous research focusing on imbalance problem, and we divide them into three families: re-sampling methods, re-weighting methods, and transfer learning based methods. The re-sampling strategies ~\cite{zhou2020bbn,wang2019dynamic,kang2019decoupling} mainly by over-sampling the minority samples and under-sampling the majority samples to construct balanced data distribution. According to the proportion of the samples, re-weighting methods  ~\cite{chou2020remix,cao2019learning,cui2019class,jamal2020rethinking} assign appropriate weights by designing re-weighted loss to balance the data distribution, whose core idea lies on the intuition that tail categories should have larger loss weights. Inspired by transferring learning, some literature ~\cite{xiang2020learning,liu2019large,liu2020deep} tries to transfer knowledge from head-classes to tail-classes to improve the diversity of the tail classes. Currently, some works focus on imbalanced node classification on graphs. ~\cite{zhao2021graphsmote} over-samples the minority class by synthesizing more natural nodes as well as relation information. ~\cite{chen2021topology} points out the unique topology-imbalance problem on graphs, and performs instance-level re-weighting based on the distance of each labeled node to its class boundary.

In this work, we mainly focus on the imbalance problems in GCN-based linkage prediction task. To the best of our knowledge, it's the first work related to imbalance problems in GCN-based linkage prediction task. The imbalance problems of GCN-based tasks lies on two sides, in addition to the number of nodes for each class, the diversity of subgraph structures are also imbalanced. As illustrated in Fig.\ref{Fig.1} (a), if a node is surrounded by too many nodes with the same identity, subgraph constructed based on the k-nearest neighbors (KNN) is extremely imbalanced. Although traditional re-sampling methods can reduce the imbalanced labels problem, it is invalid to the biased graph representations problem. Fortunately, the RIWS strategy proposed in this paper could alleviate the problems by constructing diverse subgraphs with balanced distribution.

\begin{figure*}[!htb]
  \centering
  \includegraphics[width=0.97\textwidth]{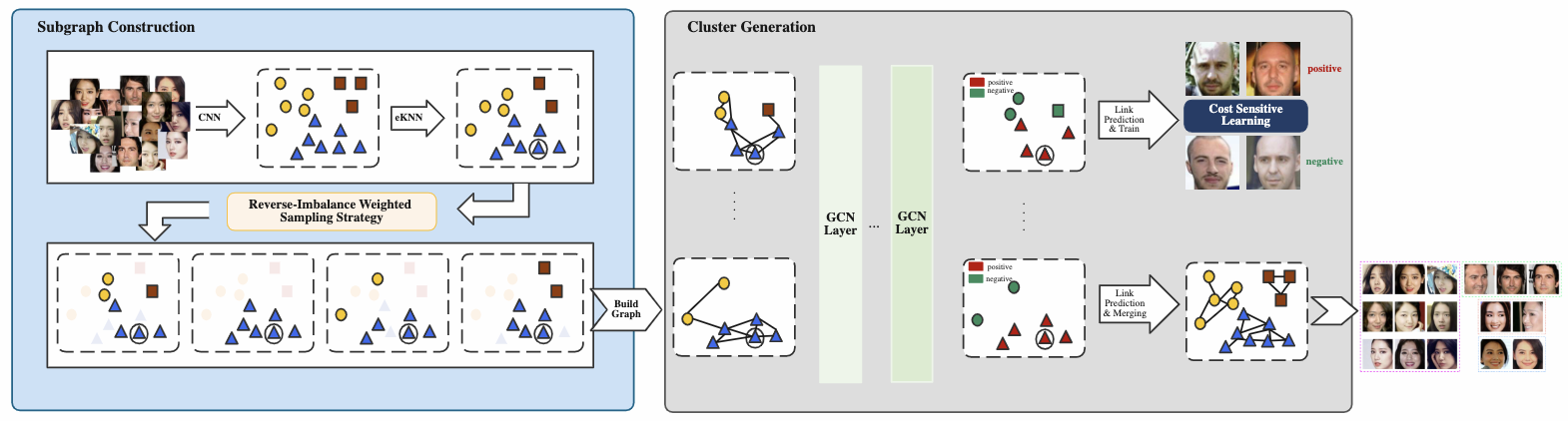}  \caption{DIGL Framework. (1) Subgraph construction stage, includes candidate nodes generation and diverse subgraph construction; (2) Cluster generation stage, contains linkage prediction and linkage merging.}
  \label{Fig.2}
\end{figure*}

\section{Feasibility of Non-graphic Methods}
This paper mainly studies the imbalance problems in GCN-based linkage prediction task. There are already numerous researchers focusing on the imbalance problem between positive and negative samples. We select some representative methods and evaluate their effectiveness in GCN-based linkage prediction task. The current mainstream methods mainly include cost sensitive learning and re-sampling methods. 
\subsection{Cost Sensitive Learning}\label{sec 3.1}
\leavevmode
\newline
Cost sensitive learning, also named as re-weighting methods. 
Both ~\cite{zhao2021graphsmote} and ~\cite{chen2021topology} are based on the node classification task to deal with the imbalance problem on the graph. From the experimental results, the class balance loss is a strong baseline, therefore, we also choose class balance loss as a comparison method. Focal loss is another widely validated re-weighting method, so we also evaluate its performance on graphs.
\subsubsection{Class balance loss}
\leavevmode
\newline
Class balance loss~\cite{cui2019class} is a strong and effective method which beats many recent proposed approached such as G-SMOTE, etc..\cite{chen2021topology,zhao2021graphsmote} It firstly calculates the average loss values for positive and negative samples respectively, and then take the average of the two loss values as the final loss value. 

\begin{center}
\begin{equation}
\begin{aligned}
\mathcal{L}_{CB}(\mathbf{z})=-\alpha_{P} y\log \left(\frac{\exp \left(z_{P}\right)}{\exp \left(z_{P}\right) + \exp \left(z_{N}\right)}\right) \\ -\alpha_{N} (1-y)\log \left(\frac{\exp \left(z_{P}\right)}{\exp \left(z_{N}\right) + \exp \left(z_{N}\right)}\right),
\end{aligned}
\end{equation}
\end{center}
where $z_P$ and $z_N$ are logit for positive sample and negative sample respectively. $\alpha_{P}$ and $\alpha_{N}$ are weights computed according to the frequency of occurrence, and they satisfy $\alpha_{P} = \frac{1}{N_p}$, $ \alpha_{N} = \frac{1}{N_N}$. ($N_P$ and $N_N$ are the number of positive and negative samples respectively).
\subsubsection{Focal loss}
\leavevmode
\newline
Focal loss was first proposed in ~\cite{lin2017focal} for object detection, and it is specially designed to handle difficult training samples. 
In our methods, we want to determine if there is an edge between pivot and its 1-hop nodes. If we define the output probability for a 1-hop node as $P=[p_P, p_N]$, where $p_P$ indicates the probability of existing an edge.
Then the focal loss for linkage prediction can be formulated as
\begin{center}
\begin{equation}
\begin{aligned}
\mathcal{L}_{\text {Focal }}(\mathbf{z})= -y\alpha_{P} (1-p_P)^{\gamma}log(p_P) \\ -(1-y)\alpha_{N}(1-p_N)^{\gamma}log(p_N).
\end{aligned}
\end{equation}
\end{center}
where $y$ is the ground truth label, and $y=1$ if there exists an edge. Hyper-parameter $\alpha$ ($\alpha_P + \alpha_N = 1$) is utilized to balance the impact of positive and negative samples, and $\gamma$ is used for mining hard examples.

\subsection{Re-sampling methods}\label{sec 3.2}

\leavevmode
\newline
The re-sampling methods mainly include random over-sampling and random under-sampling. Random over-sampling replicates random samples in minority classes, while random under-sampling randomly removes samples in majority classes. Despite simplicity, these methods can achieve good performance.

\section{Proposed Methods}
In the GCN-based linkage prediction task, the imbalanced data could cause two critical problems: imbalanced linkage labels and biased graph representations. The former is similar to the imbalance problem for image classification, that is, the imbalance between positive and negative samples. The latter is a unique problem in GCN-based tasks. The subgraphs constructed directly as L-GCN tend to be biased towards the distribution of the training set, which is prone to overfitting. To address these challenges, we propose a linkage-based Doubly Imbalanced Graph Learning (DIGL) framework for face clustering.

\subsection{Framework Overview}
\leavevmode
\newline
To better perform the clustering process on imbalanced
datasets, we decompose the problem into two aspects. One is how to build more diverse subgraphs, which can improve the generalization ability of the model. The other is how to adjust the weights to force the model to learn a balanced distribution instead of fitting the imbalanced data distribution of the train set. 


As shown in Fig.\ref{Fig.2}, our framework is mainly divided into two stages, subgraph construction and cluster generation. The former generates candidate nodes and then constructs diverse subgraphs; the latter controls the weights in the back-propagation process through the loss function, and merges the results of link prediction through the pseudo label propagation method~\cite{zhan2018consensus} to complete the clustering process.

In order to facilitate explanation, let's use $G=\{V, E\}$ to denote the feature graph, where $V= \{v_1, v_2, ..., v_N\}$ is a set of nodes in the feature space $R^{d}$.
Assuming each face's identity is represented by $Y_i$, and $N$ face images can be divided into $C$ identities $\{Y_1, Y_2, ..., Y_C\}$. In the real scenes, the ratio between positive and negative samples is extremely imbalanced, and it poses great challenge to the face clustering problems. We formulate this task as
\begin{center}
    \begin{equation}
        \hat{Y} =f(X',g(X',A'),\theta),
    \end{equation}
\end{center}
where $\hat{Y}$ is the predicted result. $X'$ and $A'$ denote the subgraph's features and adjacency matrix of sampling neighbors. g(·) is the mean aggregation operation, and $\theta$ is the learned weights.

\subsection{Subgraph Construction}
\leavevmode
\newline
In order to construct diverse subgraph samples, we first need to choose appropriate candidate nodes, and then use a specific strategy to generate subgraphs based on the candidate nodes.
\leavevmode
\newline
\subsubsection{Candidate nodes generation}
\leavevmode
\newline
Similar to L-GCN~\cite{wang2019linkage}, we take each instance in the dataset as a center (called ``pivot'') and a subgraph is constructed for each pivot. Then we estimate the linkage likelihood between two nodes in a graph. Since L-GCN does not consider the imbalance problem, compared with the initial subgraph construction method, we adds an expansion coefficient $\gamma$. With this coefficient, the selection interval of 1-hop nodes is increased from k to k*$\gamma$, which is defined as expanded k nearest neighbors (eKNN) and then k nodes in eKNN are selected as 1-hop neighbors, which is used to control the positive and negative samples' distribution to a certain extent. As shown in Fig.~\ref{Fig.1} (a), the blue center point surrounded by the solid green circle is the pivot, and the k*$\gamma$ nodes closest to the pivot in Fig.~\ref{Fig.1} (a) form its eKNN (where k=10, $\gamma$=1.5). The selection of 2-hop nodes and the edge connection process are consistent with L-GCN.\\
\subsubsection{Diverse Subgraph Construction}
\leavevmode
\newline
Existing works did not consider the imbalance problem in the task of GCN-based face clustering. Although there are a lot of works devoted to the imbalance problem, however, the community only takes the imbalanced linkage labels into account, without realizing the biased graph representation problem of GCNs. In this section, we first introduce the representative re-sampling method (random over-sampling and under-sampling) to alleviate the imbalance problem of positive and negative samples in GCNs. Furthermore, in order to simultaneously consider the biased graph representation problem, we propose a Reverse-Imbalance Weighted Sampling (RIWS) strategy, which can effectively improve the diversity of the subgraph structure and alleviate the biased subgraph representation problem.

The re-sampling method is mainly to build a balanced subgraph for each pivot by over-sampling tail categories and under-sampling head categories for its candidate nodes. Although the construction result is balanced from the perspective of each single subgraph, but the diversity of subgraph structures will be relatively poor. The generalization ability may be limited when faced with test data with complex distribution in real scenarios.

The procedure of proposed RIWS method is as follows. Firstly, for each pivot, the nodes in its eKNN are selected as candidate nodes of its subgraph, which is explained in candidate nodes generation section. Secondly, the weight of each candidate sample is calculated according to the number of positive and negative samples by 

\begin{equation}
w_{i}^{j}=
\begin{cases}
\frac{1}{2}*\frac{1}{\textstyle{} \sum_{v_{j} \in N _{i}}\left(\mathbbm{1}_{y_{j} = y_{i}}\right)} 
& \text{if $y_{i}$ = $y_{j}$}\\
\frac{1}{2}*\frac{1}{\textstyle{} \sum_{v_{j} \in N _{i}}\left(\mathbbm{1}_{y_{j} \ne y_{i}}\right)} 
& \text{if $y_{i}$ $\ne$ $y_{j}$}
\end{cases},
\end{equation}
where $w_{i}^{j}$ is the weight of the j-th neighbour node in the i-th sample's eKNN (i.e., the probability of j-th neighbour node to be selected as a node in the subgraph pivoted by the i-th sample node). Then, k nodes are selected from each eKNN based on these weights to construct its subgraph. The node feature normalization and subgraph edge connection process are similar to L-GCN.

Fig.\ref{Fig.1} demonstrates the selected 1-hop nodes of some subgraph examples by different methods. Assuming k=10 and $\gamma$=1.5 in each sub-picture, the node surrounded by a green solid circle is the pivot sample, and the 1-hop candidate nodes are composed of the pivot's eKNN in the feature space, of which the nodes in same category as the pivot are positive samples, otherwise they are negative samples. Fig.\ref{Fig.1} (a) indicates the selected k neighbors with L-GCN, where 10 nodes closest to the pivot in feature space are selected, without considering the imbalance problem of samples. When conventional re-sampling strategy is adopted, the selected 1-hop neighbors is illustrated in Fig.\ref{Fig.1} (b), where 5 positive samples (blue dots) are randomly selected (under-sampling), and 5 negative samples are selected (over-sampling). If the number of negative samples is insufficient, over-sampling needs to be carried out by duplication. While using the RIWS strategy, the balance and diversity of subgraphs can both be guaranteed. On one hand, the balancing weight can ensure that the overall distribution of positive and negative samples in the 1-hop neighbors is balanced; on the other hand, the weighted random sampling process provides diverse structures (including all those shown in Fig.\ref{Fig.1} (a, b, c, d), and more). These two properties together help address the problems of both imbalanced labels and biased graph representations.

\subsection{Cluster Generation}
\leavevmode
\newline
After the subgraphs are constructed, the clustering results will be obtained through the clustering generation stage. 
\leavevmode
\newline
\subsubsection{Linkage Prediction}
\leavevmode
\newline
The context of each pivot (nodes and edges in the subgraph) contains rich information, which is of great significance for judging whether the pivot and the nodes in its subgraph belong to the same category (existing edge links). To exploit this information, we employ an graph convolution like L-GCN to propagate the subgraphs. The graph convolutional layer takes feature matrix X and adjacency matrix A as input, and takes transformed feature matrix Y as output. The input feature matrix of the first layer is the original face images' features, and the graph convolution of each layer is formulated as
\begin{center}
    \begin{equation}
        Y=\sigma ([X||g(X,A)]W),
    \end{equation}
\end{center}
where g(·) is the mean aggregation operation, $W$ is the learnable weights, and $\parallel $ represents feature matrix concatenation operation.

Whether the linkage exists between the pivot and its 1-hop neighbors is a binary classification problem. If the model is trained on a imbalanced dataset, the pivots' KNN may be dominated by the majority positive or negative samples. Taking Fig.~\ref{Fig.1} (a) as an example, the pivot' KNN is dominated by positive samples, which leads to a imbalanced subgraph, and then seriously affect the learning of the model. Different from L-GCN, we introduce re-weighting methods (focal loss and class balance loss, which is explained in \ref{sec 3.1}) to balance the weights of positive and negative samples in each subgraph.
\leavevmode
\newline
\subsubsection{Linkage Merging}
\leavevmode
\newline
Similar to L-GCN, we construct diversified subgraphs with each sample as a pivot, and then predict whether there is a link between the pivot and each node in its subgraph. Further, we deal with the imbalance problem from two levels, improving the generalization ability of the model. In this way, we get a series of edge link relations, which we can think of as a complete graph consisting of all nodes. Finally, since the clustering results are very sensitive to a single threshold, we complete the clustering process through the pseudo label propagation strategy proposed in ~\cite{zhan2018consensus}. Specifically, we iteratively process the complete graph according to the following steps, (1) define a smaller threshold thresh and a $max\_size$ parameter; (2) delete the edges with weight smaller than the threshold in the graph, and take out the cluster smaller than $max\_size$ as partial results; (3) for clusters larger than $max\_size$, we increase the thresh parameter by a specific step, (4) repeat step 2 and 3 until all results are obtained. Thus we could obtain the final clustering result.

\section{Experiments}

In this section, we construct imbalanced datasets and conduct extensive experiments with the goal to answer the following research questions. 
	
	\smallskip\noindent\textbf{Q1:} Are the methods for conventional imbalanced classification problem still effective when extended to GCN-based linkage prediction task? Does DIGL outperform the previous methods?
	
	\smallskip\noindent\textbf{Q2:} Does DIGL own good generalization ability and whether the model can be effective in different datasets and different domain?

	\smallskip\noindent\textbf{Q3:} What is the effect of hyper-parameters of re-sampling and RIWS methods in DIGL?


\linespread{1.16}
\begin{table*}[!htb]
    \caption{PR AUC on MS-Celeb-1M imbalanced sub-datasets for each method and their combinations.}
    \centering
    \resizebox{\linewidth}{!}{
    \begin{tabular}{lccccccccc}
    \toprule
        Method& (200,3)&  (200,5)&  (500,3)&  (500,5)&  (1000,3)&  (1000,5)&  (2000,3)&  (2000,5)&  Avg\\ \hline \hline
        L-GCN&  0.9588&  0.9686&  0.9606&  0.9683&  0.9672&  0.9716&  0.9772&  0.9778&  0.9688 \\
        RS&  0.9625&  0.9679&  0.9698&  0.9754&  0.9758&  0.9770&  0.9819&  0.9809&  0.9739   \\ 
        RIWS&  0.9658&  0.9729&  0.9739&  0.9772&  0.9813&  0.9825&  0.9860&  0.9867&  0.9783    \\ \hline \hline
        
        CB&  0.9660&  0.9731&  0.9683&  0.9733&  0.9746&  0.9769&  0.9817&  0.9823&  0.9745    \\ 
        CB (w/ RS)&  0.9681&  0.9697&  0.9736&  0.9763&  0.9762&  0.9782&  0.9833&  0.9839&  0.9762   \\
        CB (w/ RIWS)&  0.9699&  0.9754&  0.9774&  0.9786&  0.9822&  0.9826&  \textbf{0.9868}&  \textbf{0.9875}&  0.9800   \\ \hline \hline
        
        FL&  0.9694&  0.9731&  0.9720&  0.9745&  0.9760&  0.9763&  0.9816&  0.9805&  0.9754   \\ 
        FL (w/ RS)&  \textbf{0.9740}&  0.9761&  0.9777&  0.9788&  0.9795&  0.9804&  0.9804&  0.9815&  0.9786   \\
        FL (w/ RIWS)&  0.9736&  \textbf{0.9764}&  \textbf{0.9791}&  \textbf{0.9800}&  \textbf{0.9828}&  \textbf{0.9830}&  0.9858&  0.9858&  \textbf{0.9808}  \\ 
    \bottomrule
    \end{tabular}
    }
    \label{Table.1}
\end{table*}

\subsection{Settings}
\subsubsection{Imbalanced datasets construction}
\leavevmode
\newline
In order to evaluate the performance of each method on imbalanced
datasets, referring to~\cite{liu2020deep}, we construct a series of imbalanced
datasets based on two public datasets: MS-Celeb-1M~\cite{guo2016ms} and DeepFashion~\cite{liu2016deepfashion}. Taking MS-Celeb-1M as an example, the construction procedure of the imbalanced datasets is as follows.

Based on part0 of the cleaned MS-Celeb-1M~\cite{yang2020learninggcnve}, we
synthesized 8 imbalanced training sets according to two
hyper-parameters: majority\_identity\_count $m$ and
minority\_identity\_size $n$. Specifically, the identities are sorted by their
number of samples, and top $m$ identities are selected as the majority classes.
For the remaining part, $n$ samples are taken randomly from each identity. If the
identity size is lower than $n$, all samples would be taken. We adopt 200, 500,
1000, 2000 for $m$, and 3, 5 for $n$. In this way, 8 imbalanced datasets can be
constructed, denoted as (H200, S3), (H200, S5), and so on.
We train models on 8 imbalanced datasets, and then test them on part1 of the
cleaned MS-Celeb-1M, respectively.

Similar to MS-Celeb-1M, we also constructed 2 imbalanced training set based on DeepFashion.
\subsubsection{Evaluation metrics}
\leavevmode
\newline
Since linkage merging stage is a heuristic process and contains several hyper-parameters, so we decouple edge classification and linkage merging to eliminate the influence of hyper-parameters tuning. In the stage of imbalanced edge classification, we select PR AUC (AUC of Precision and Recall Curve) as the evaluation metric rather than ROC AUC, because in the case of highly skewed data sets, it is observed that the ROC curve may provide an overly optimistic view of an algorithm’s performance, while the PR curves can provide a more informative representation of performance assessment under such situations \cite{he2009learning}. As for clustering stage, Bcubed F score is selected. 

\subsection{Method Comparison}
\leavevmode
\newline
In this section, we investigate the performance of each method and their combinations, and PR AUC of edge classification is selected as the basic metric.

Table \ref{Table.1} shows the edge classification PR AUC of L-GCN and other methods on the constructed 8 datasets based on MS-Celeb-1M. Among these methods, 
CB, FL, RS stands for class balance loss, focal loss and conventional re-sampling method respectively.
The four methods can further be divided into two categories: re-sampling methods and re-weighting methods. The former methods (i.e. RS and RIWS) are mainly used in the subgraph construction stage, while the latter (i.e. CB and FL) are applied for the training of the classifier.

\subsubsection{Comparison with each competitor.}
\leavevmode
\newline
Obviously, all methods achieve better performance than baseline named L-GCN except for individual case,  noticing that the imbalance problem which is ignored by previous method should be carefully addressed.

In the sub-datasets with a smaller majority identity count ( i.e. sub-dataset H200-S3 and H200-S5 ), re-weighting methods achieve better performacne than re-sampling methods. We suspected that it's because imbalanced ratio problem between positive and negative samples is serious in this configuration, while the sampling-based methods perform poorly due to lack of sufficient samples to be sampled.  Focal loss (FL) gets more significant improvements than class balance loss (CB), because focal loss (FL) assigns different weights to different samples and alleviate the imbalance problem by its ability of learning on hard examples and down-weight the numerous easy negatives. However, class balance loss (CB) gives uniform weight to samples of the same category, leading to inferior performance.
 
 In the sub-dataset with a larger majority identity count, the effect of the biased graph representations is gradually manifested. The RIWS method proposed in this paper achieves better results than all other methods, which has reached a high AUC of 0.9867 in sub-dataset (H2000,S5), demonstrating that the biased graph representations problem is successfully alleviated by RIWS.
\newline

\subsubsection{Study on the combinations of re-sampling and re-weighting method.}
\leavevmode
\newline
In order to further demonstrate the performance of each method, we combine the two categories of methods (i.e. re-sampling methods and re-weighting methods), and the experimental results are also shown in the table \ref{Table.1}. Based on the experiment results, we have two findings. First, we can observe that the effectiveness of the combinations of the two categories of methods are better than that of the single method, demonstrating that it can complement each other to solve the imbalance problem from both subgraph construction stage and classifier stage. Second, we can also find that whether RIWS combined with CB or FL, achieve consistent promotions across all datasets than RS to combined with them, which once again proves the effectiveness of our proposed method, RIWS. 


\subsection{Generalization Analysis}
\leavevmode
\newline
The generalization of a method is an important indicator to measure the performance of a method, we will discuss the generalization of RIWS from two aspects: (1) generalization on different data domain;
(2) generalization on different data distribution.

\begin{table}[]
\caption{Performance on DeepFashion imbalanced sub-datasets.}
\centering
\resizebox{\linewidth}{!}{
\begin{tabular}{l|cc|cc}
\hline
 & \multicolumn{2}{c|}{(H200,S3)} & \multicolumn{2}{c}{(H500,S3)} \\ \cline{2-5} 
    & PR AUC  & F score & PR AUC  & F score \\ \hline
L-GCN & 0.7897 & 0.5769     & 0.7981 & 0.5918 \\
RS & 0.7917 & 0.5818     & 0.7962 & 0.5853     \\
RIWS & 0.7933 & 0.5879     & 0.8002 & 0.5952     \\\hline
FL & 0.7643 & 0.5448     & 0.7646 & 0.5179     \\\hline
CB & 0.7962 & 0.5894     & 0.8062 & 0.5974     \\
CB (w/ RS) & 0.7938 & 0.5827     & 0.8049 & 0.5944     \\
CB (w/ RIWS) & \textbf{0.7981} & \textbf{0.5942}     & \textbf{0.8066} & \textbf{0.6001}     \\ \hline
\end{tabular}
}
\label{table.2}
\end{table}

\begin{table}[]
\caption{Comparison with baseline trained on full dataset.}
\centering
\setlength{\tabcolsep}{1mm}
\resizebox{\linewidth}{!}{
\begin{tabular}{l|c|c|cc}
\hline
Method  & training set        & samples & PR AUC     & F score \\ \hline
L-GCN   & MS-Celeb-1M    & 576k    & 0.9839 & 0.8437* \\
L-GCN   & (H2000, S3)    & 250k    & 0.9772 & 0.7924 \\
FL (w/ RIWS) & (H2000, S3) & 250k    & 0.9858 & 0.8397  \\ \hline
L-GCN   & DeepFashion    & 26k    & 0.8076 & 0.6013* \\
L-GCN & (H500,S3) & 19k    & 0.7981 & 0.5918  \\
CB (w/ RIWS) & (H500,S3) & 19k    & 0.8066 & 0.6001  \\ \hline
\end{tabular}
}
\label{table.3}
\end{table}

\subsubsection{Generalization on Data Domain}
\leavevmode
\newline
In order to validate the generalization ability of the methods on different data domain, we conduct experiments on two sub-datasets of DeepFashion (i.e. Fashion Clustering data), which is (H200, S3) and (H500,S3). As shown in table \ref{table.2}, all methods except focal loss exceed L-GCN baseline, and class balance loss combined with RIWS achieves the best performance, indicating that RIWS owns well domain generalization ability.

\subsubsection{Generalization on Data Distribution}
\leavevmode
\newline
Table \ref{table.3} demonstrates the comparison between the best combination of imbalanced methods and the baseline. Notice that our combination is trained on imbalanced sub-dataset, while baseline is trained on the full dataset. Results marked with an asterisk are obtained from \cite{yang2020learninggcnve}. Although trained on less data with imbalanced distribution, our method obtains comparable results with baseline trained on full MS-Celeb-1M or DeepFashion datasets, which strongly validates the effectiveness of our methods. 
\subsection{Hyper-parameter sensitivity. }
\begin{figure}[!htp]
  \centering
  \setlength{\belowcaptionskip}{0cm}
  \setlength{\abovecaptionskip}{0cm}
  \includegraphics[width=0.5\textwidth]{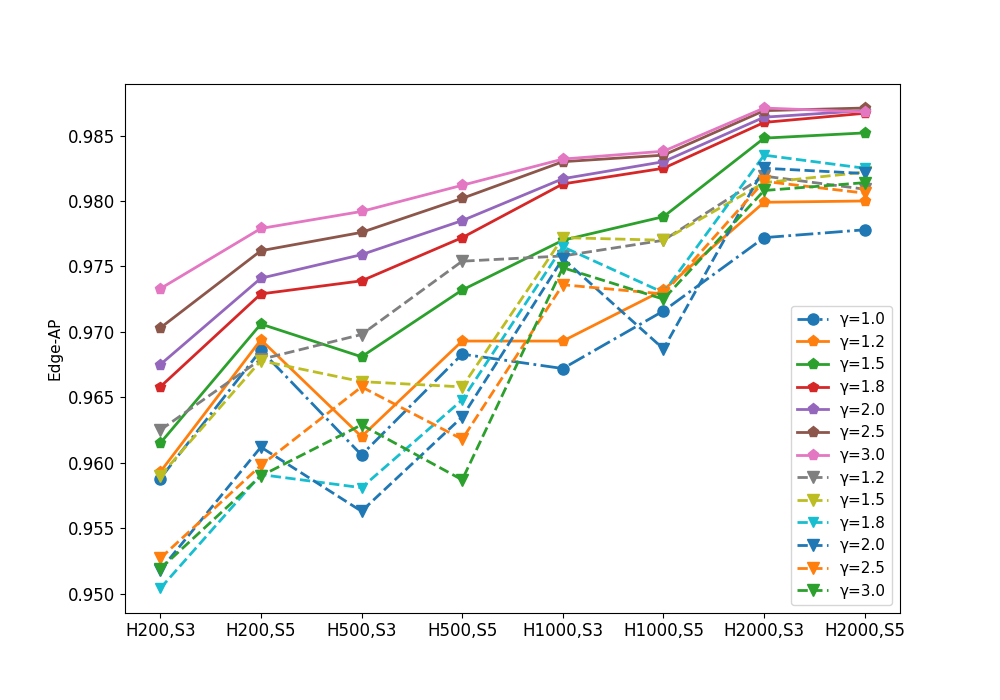}
  \caption{Hyper-parameter sensitivity comparison between re-sampling and RIWS Methods.}
  \label{Fig.3}
\end{figure}

\leavevmode
\newline
In this paper, we introduce a new parameter $\gamma$, which controls the ratio of candidate nodes and selected neighbors of each subgraph. When $\gamma=1$, our subgraph construction method is equivalent to the L-GCN baseline. The performance with different $\gamma$ values of regular re-sampling and RIWS methods are shown in Fig.\ref{Fig.3}. The dot-and-dash line, the dashed line and the solid line denote respectively baseline, re-sampling and RIWS method. And each method alleviates the imbalance problems based on the baseline.

As $\gamma$ increases, the performance of the re-sampling method first increases and then decreases, and reaches the maximum value at $\gamma$=1.2, while RIWS continues to increase as the growth of $\gamma$, with the growth rate beginning to slow down sharply at $\gamma$=2.0. In our experiments, we select $\gamma$=1.2 for re-sampling method, and in order not to add too much computational overhead, 2.0 is chosen for the RIWS method. In this configuration, re-sampling and RIWS methods obtain average PR AUC values on 8 imbalanced sub-datasets of 0.9739 and 0.9783, respectively, which both far exceed the baseline of 0.9688. And our proposed RIWS method is significantly better than regular re-sampling method in each sub-dataset.

\section{Conclusion and Discussion}

The imbalance problems in GCN-based linkage prediction tasks were studied in this paper for the first time from two aspects: imbalanced labels and biased graph representations.
Extensive experiments were conducted to evaluate the effectiveness of four typical approaches to the imbalance problem in image classification on addressing the imbalanced label problem in GCN-based tasks, showing that all of them can bring some extent of performance improvement, and that some of their combinations can further extend the improvement.
A Reverse-Imbalance Weighted Sampling (RIWS) strategy was proposed in this paper as a trial to tackle the problems of both imbalanced labels and biased graph representations, whose effectiveness was demonstrated in numerous experiments on MS-Celeb-1M and DeepFashion datasets.

These results provide some references on selecting and designing approaches to tackling imbalance problems in GCN-based node and edge classification tasks where imbalanced data can cause problems in the aforementioned two aspects: imbalanced labels and biased graph representations. The RIWS strategy is a trial of designing approaches to addressing the two problems simultaneously, but not necessarily the optimal one. More studies are required and welcomed for better solutions for various scenarios. 

\bibliography{reference}

\begin{thebibliography}{10}
\providecommand{\url}[1]{#1}
\csname url@samestyle\endcsname
\providecommand{\newblock}{\relax}
\providecommand{\bibinfo}[2]{#2}
\providecommand{\BIBentrySTDinterwordspacing}{\spaceskip=0pt\relax}
\providecommand{\BIBentryALTinterwordstretchfactor}{4}
\providecommand{\BIBentryALTinterwordspacing}{\spaceskip=\fontdimen2\font plus
\BIBentryALTinterwordstretchfactor\fontdimen3\font minus
  \fontdimen4\font\relax}
\providecommand{\BIBforeignlanguage}[2]{{%
\expandafter\ifx\csname l@#1\endcsname\relax
\typeout{** WARNING: IEEEtran.bst: No hyphenation pattern has been}%
\typeout{** loaded for the language `#1'. Using the pattern for}%
\typeout{** the default language instead.}%
\else
\language=\csname l@#1\endcsname
\fi
#2}}
\providecommand{\BIBdecl}{\relax}
\BIBdecl

\bibitem{wang2019linkage}
Z.~Wang, L.~Zheng, Y.~Li, and S.~Wang, ``Linkage based face clustering via
  graph convolution network,'' in \emph{Proceedings of the IEEE/CVF Conference
  on Computer Vision and Pattern Recognition}, 2019, pp. 1117--1125.

\bibitem{yang2019learninggcnds}
L.~Yang, X.~Zhan, D.~Chen, J.~Yan, C.~C. Loy, and D.~Lin, ``Learning to cluster
  faces on an affinity graph,'' in \emph{Proceedings of the IEEE/CVF Conference
  on Computer Vision and Pattern Recognition}, 2019, pp. 2298--2306.

\bibitem{yang2020learninggcnve}
L.~Yang, D.~Chen, X.~Zhan, R.~Zhao, C.~C. Loy, and D.~Lin, ``Learning to
  cluster faces via confidence and connectivity estimation,'' in
  \emph{Proceedings of the IEEE/CVF Conference on Computer Vision and Pattern
  Recognition}, 2020, pp. 13\,369--13\,378.

\bibitem{guo2020density}
S.~Guo, J.~Xu, D.~Chen, C.~Zhang, X.~Wang, and R.~Zhao, ``Density-aware feature
  embedding for face clustering,'' in \emph{Proceedings of the IEEE/CVF
  Conference on Computer Vision and Pattern Recognition}, 2020, pp. 6698--6706.

\bibitem{lloyd1982leastkmeans}
S.~Lloyd, ``Least squares quantization in pcm,'' \emph{IEEE Transactions on
  Information Theory}, vol.~28, no.~2, pp. 129--137, 1982.

\bibitem{ester1996densitydbscan}
M.~Ester, H.-P. Kriegel, J.~Sander, X.~Xu \emph{et~al.}, ``A density-based
  algorithm for discovering clusters in large spatial databases with noise.''
  in \emph{Kdd}, vol.~96, no.~34, 1996, pp. 226--231.

\bibitem{sibson1973slinkhac}
R.~Sibson, ``Slink: an optimally efficient algorithm for the single-link
  cluster method,'' \emph{The Computer Journal}, vol.~16, no.~1, pp. 30--34,
  1973.

\bibitem{wang2021contrastive}
P.~Wang, K.~Han, X.-S. Wei, L.~Zhang, and L.~Wang, ``Contrastive learning based
  hybrid networks for long-tailed image classification,'' \emph{arXiv preprint
  arXiv:2103.14267}, 2021.

\bibitem{zhou2020bbn}
B.~Zhou, Q.~Cui, X.-S. Wei, and Z.-M. Chen, ``Bbn: Bilateral-branch network
  with cumulative learning for long-tailed visual recognition,'' in
  \emph{Proceedings of the IEEE/CVF Conference on Computer Vision and Pattern
  Recognition}, 2020, pp. 9719--9728.

\bibitem{kang2019decoupling}
B.~Kang, S.~Xie, M.~Rohrbach, Z.~Yan, A.~Gordo, J.~Feng, and Y.~Kalantidis,
  ``Decoupling representation and classifier for long-tailed recognition,''
  \emph{arXiv preprint arXiv:1910.09217}, 2019.

\bibitem{wang2019dynamic}
Y.~Wang, W.~Gan, J.~Yang, W.~Wu, and J.~Yan, ``Dynamic curriculum learning for
  imbalanced data classification,'' in \emph{Proceedings of the IEEE/CVF
  International Conference on Computer Vision}, 2019, pp. 5017--5026.

\bibitem{cui2019class}
Y.~Cui, M.~Jia, T.-Y. Lin, Y.~Song, and S.~Belongie, ``Class-balanced loss
  based on effective number of samples,'' in \emph{Proceedings of the IEEE/CVF
  Conference on Computer Vision and Pattern Recognition}, 2019, pp. 9268--9277.

\bibitem{cao2019learning}
K.~Cao, C.~Wei, A.~Gaidon, N.~Arechiga, and T.~Ma, ``Learning imbalanced
  datasets with label-distribution-aware margin loss,'' \emph{arXiv preprint
  arXiv:1906.07413}, 2019.

\bibitem{jamal2020rethinking}
M.~A. Jamal, M.~Brown, M.-H. Yang, L.~Wang, and B.~Gong, ``Rethinking
  class-balanced methods for long-tailed visual recognition from a domain
  adaptation perspective,'' in \emph{Proceedings of the IEEE/CVF Conference on
  Computer Vision and Pattern Recognition}, 2020, pp. 7610--7619.

\bibitem{chou2020remix}
H.-P. Chou, S.-C. Chang, J.-Y. Pan, W.~Wei, and D.-C. Juan, ``Remix: Rebalanced
  mixup,'' in \emph{European Conference on Computer Vision}.\hskip 1em plus
  0.5em minus 0.4em\relax Springer, 2020, pp. 95--110.

\bibitem{liu2020deep}
J.~Liu, Y.~Sun, C.~Han, Z.~Dou, and W.~Li, ``Deep representation learning on
  long-tailed data: A learnable embedding augmentation perspective,'' in
  \emph{Proceedings of the IEEE/CVF Conference on Computer Vision and Pattern
  Recognition}, 2020, pp. 2970--2979.

\bibitem{xiang2020learning}
L.~Xiang, G.~Ding, and J.~Han, ``Learning from multiple experts: Self-paced
  knowledge distillation for long-tailed classification,'' in \emph{European
  Conference on Computer Vision}.\hskip 1em plus 0.5em minus 0.4em\relax
  Springer, 2020, pp. 247--263.

\bibitem{liu2019large}
Z.~Liu, Z.~Miao, X.~Zhan, J.~Wang, B.~Gong, and S.~X. Yu, ``Large-scale
  long-tailed recognition in an open world,'' in \emph{Proceedings of the
  IEEE/CVF Conference on Computer Vision and Pattern Recognition}, 2019, pp.
  2537--2546.

\bibitem{deng2009imagenet}
J.~Deng, W.~Dong, R.~Socher, L.-J. Li, K.~Li, and L.~Fei-Fei, ``Imagenet: A
  large-scale hierarchical image database,'' in \emph{Proceedings of the
  IEEE/CVF Conference on Computer Vision and Pattern Recognition}, 2009, pp.
  248--255.

\bibitem{guo2016ms}
Y.~Guo, L.~Zhang, Y.~Hu, X.~He, and J.~Gao, ``Ms-celeb-1m: A dataset and
  benchmark for large-scale face recognition,'' in \emph{European conference on
  computer vision}.\hskip 1em plus 0.5em minus 0.4em\relax Springer, 2016, pp.
  87--102.

\bibitem{krizhevsky2009learning}
A.~Krizhevsky, G.~Hinton \emph{et~al.}, ``Learning multiple layers of features
  from tiny images,'' \emph{Tech Report}, 2009.

\bibitem{lin2017focal}
T.-Y. Lin, P.~Goyal, R.~Girshick, K.~He, and P.~Doll{\'a}r, ``Focal loss for
  dense object detection,'' in \emph{Proceedings of the IEEE international
  conference on computer vision}, 2017, pp. 2980--2988.

\bibitem{liu2016deepfashion}
Z.~Liu, P.~Luo, S.~Qiu, X.~Wang, and X.~Tang, ``Deepfashion: Powering robust
  clothes recognition and retrieval with rich annotations,'' in
  \emph{Proceedings of the IEEE conference on computer vision and pattern
  recognition}, 2016, pp. 1096--1104.

\bibitem{zhan2018consensus}
X.~Zhan, Z.~Liu, J.~Yan, D.~Lin, and C.~C. Loy, ``Consensus-driven propagation
  in massive unlabeled data for face recognition,'' in \emph{proceedings of the
  European Conference on Computer Vision (ECCV)}, 2018, pp. 568--583.

\bibitem{zhao2021graphsmote}
T.~Zhao, X.~Zhang, and S.~Wang, ``Graphsmote: Imbalanced node classification on
  graphs with graph neural networks,'' in \emph{Proceedings of the 14th ACM
  international conference on web search and data mining}, 2021, pp. 833--841.

\bibitem{chen2021topology}
D.~Chen, Y.~Lin, G.~Zhao, X.~Ren, P.~Li, J.~Zhou, and X.~Sun,
  ``Topology-imbalance learning for semi-supervised node classification,''
  \emph{Advances in Neural Information Processing Systems}, vol.~34, pp.
  29\,885--29\,897, 2021.

\bibitem{wang2022ada}
Y.~Wang, Y.~Zhang, F.~Zhang, S.~Wang, M.~Lin, Y.~Zhang, and X.~Sun, ``Ada-nets:
  Face clustering via adaptive neighbour discovery in the structure space,''
  \emph{arXiv preprint arXiv:2202.03800}, 2022.

\bibitem{he2009learning}
H.~He and E.~A. Garcia, ``Learning from imbalanced data,'' \emph{IEEE
  Transactions on knowledge and data engineering}, vol.~21, no.~9, pp.
  1263--1284, 2009.

\end{thebibliography}


\begin{thebibliography}{99}













\bibitem{wang2019linkage}
Z.~Wang, L.~Zheng, Y.~Li, and S.~Wang, ``Linkage based face clustering via
  graph convolution network,'' in \emph{Proceedings of the IEEE/CVF Conference
  on Computer Vision and Pattern Recognition}, 2019, pp. 1117--1125.

\bibitem{yang2019learninggcnds}
L.~Yang, X.~Zhan, D.~Chen, J.~Yan, C.~C. Loy, and D.~Lin, ``Learning to cluster
  faces on an affinity graph,'' in \emph{Proceedings of the IEEE/CVF Conference
  on Computer Vision and Pattern Recognition}, 2019, pp. 2298--2306.

\bibitem{yang2020learninggcnve}
L.~Yang, D.~Chen, X.~Zhan, R.~Zhao, C.~C. Loy, and D.~Lin, ``Learning to
  cluster faces via confidence and connectivity estimation,'' in
  \emph{Proceedings of the IEEE/CVF Conference on Computer Vision and Pattern
  Recognition}, 2020, pp. 13\,369--13\,378.

\bibitem{guo2020density}
S.~Guo, J.~Xu, D.~Chen, C.~Zhang, X.~Wang, and R.~Zhao, ``Density-aware feature
  embedding for face clustering,'' in \emph{Proceedings of the IEEE/CVF
  Conference on Computer Vision and Pattern Recognition}, 2020, pp. 6698--6706.

\bibitem{lloyd1982leastkmeans}
S.~Lloyd, ``Least squares quantization in pcm,'' \emph{IEEE Transactions on
  Information Theory}, vol.~28, no.~2, pp. 129--137, 1982.

\bibitem{ester1996densitydbscan}
M.~Ester, H.-P. Kriegel, J.~Sander, X.~Xu \emph{et~al.}, ``A density-based
  algorithm for discovering clusters in large spatial databases with noise.''
  in \emph{Kdd}, vol.~96, no.~34, 1996, pp. 226--231.

\bibitem{sibson1973slinkhac}
R.~Sibson, ``Slink: an optimally efficient algorithm for the single-link
  cluster method,'' \emph{The Computer Journal}, vol.~16, no.~1, pp. 30--34,
  1973.

\bibitem{wang2021contrastive}
P.~Wang, K.~Han, X.-S. Wei, L.~Zhang, and L.~Wang, ``Contrastive learning based
  hybrid networks for long-tailed image classification,'' \emph{arXiv preprint
  arXiv:2103.14267}, 2021.

\bibitem{zhou2020bbn}
B.~Zhou, Q.~Cui, X.-S. Wei, and Z.-M. Chen, ``Bbn: Bilateral-branch network
  with cumulative learning for long-tailed visual recognition,'' in
  \emph{Proceedings of the IEEE/CVF Conference on Computer Vision and Pattern
  Recognition}, 2020, pp. 9719--9728.

\bibitem{kang2019decoupling}
B.~Kang, S.~Xie, M.~Rohrbach, Z.~Yan, A.~Gordo, J.~Feng, and Y.~Kalantidis,
  ``Decoupling representation and classifier for long-tailed recognition,''
  \emph{arXiv preprint arXiv:1910.09217}, 2019.

\bibitem{wang2019dynamic}
Y.~Wang, W.~Gan, J.~Yang, W.~Wu, and J.~Yan, ``Dynamic curriculum learning for
  imbalanced data classification,'' in \emph{Proceedings of the IEEE/CVF
  International Conference on Computer Vision}, 2019, pp. 5017--5026.

\bibitem{cui2019class}
Y.~Cui, M.~Jia, T.-Y. Lin, Y.~Song, and S.~Belongie, ``Class-balanced loss
  based on effective number of samples,'' in \emph{Proceedings of the IEEE/CVF
  Conference on Computer Vision and Pattern Recognition}, 2019, pp. 9268--9277.

\bibitem{cao2019learning}
K.~Cao, C.~Wei, A.~Gaidon, N.~Arechiga, and T.~Ma, ``Learning imbalanced
  datasets with label-distribution-aware margin loss,'' \emph{arXiv preprint
  arXiv:1906.07413}, 2019.

\bibitem{jamal2020rethinking}
M.~A. Jamal, M.~Brown, M.-H. Yang, L.~Wang, and B.~Gong, ``Rethinking
  class-balanced methods for long-tailed visual recognition from a domain
  adaptation perspective,'' in \emph{Proceedings of the IEEE/CVF Conference on
  Computer Vision and Pattern Recognition}, 2020, pp. 7610--7619.

\bibitem{chou2020remix}
H.-P. Chou, S.-C. Chang, J.-Y. Pan, W.~Wei, and D.-C. Juan, ``Remix: Rebalanced
  mixup,'' in \emph{European Conference on Computer Vision}.\hskip 1em plus
  0.5em minus 0.4em\relax Springer, 2020, pp. 95--110.

\bibitem{liu2020deep}
J.~Liu, Y.~Sun, C.~Han, Z.~Dou, and W.~Li, ``Deep representation learning on
  long-tailed data: A learnable embedding augmentation perspective,'' in
  \emph{Proceedings of the IEEE/CVF Conference on Computer Vision and Pattern
  Recognition}, 2020, pp. 2970--2979.

\bibitem{xiang2020learning}
L.~Xiang, G.~Ding, and J.~Han, ``Learning from multiple experts: Self-paced
  knowledge distillation for long-tailed classification,'' in \emph{European
  Conference on Computer Vision}.\hskip 1em plus 0.5em minus 0.4em\relax
  Springer, 2020, pp. 247--263.

\bibitem{liu2019large}
Z.~Liu, Z.~Miao, X.~Zhan, J.~Wang, B.~Gong, and S.~X. Yu, ``Large-scale
  long-tailed recognition in an open world,'' in \emph{Proceedings of the
  IEEE/CVF Conference on Computer Vision and Pattern Recognition}, 2019, pp.
  2537--2546.


\bibitem{guo2016ms}
Y.~Guo, L.~Zhang, Y.~Hu, X.~He, and J.~Gao, ``Ms-celeb-1m: A dataset and
  benchmark for large-scale face recognition,'' in \emph{European conference on
  computer vision}.\hskip 1em plus 0.5em minus 0.4em\relax Springer, 2016, pp.
  87--102.


\bibitem{lin2017focal}
T.-Y. Lin, P.~Goyal, R.~Girshick, K.~He, and P.~Doll{\'a}r, ``Focal loss for
  dense object detection,'' in \emph{Proceedings of the IEEE international
  conference on computer vision}, 2017, pp. 2980--2988.

\bibitem{liu2016deepfashion}
Z.~Liu, P.~Luo, S.~Qiu, X.~Wang, and X.~Tang, ``Deepfashion: Powering robust
  clothes recognition and retrieval with rich annotations,'' in
  \emph{Proceedings of the IEEE conference on computer vision and pattern
  recognition}, 2016, pp. 1096--1104.

\bibitem{zhan2018consensus}
X.~Zhan, Z.~Liu, J.~Yan, D.~Lin, and C.~C. Loy, ``Consensus-driven propagation
  in massive unlabeled data for face recognition,'' in \emph{proceedings of the
  European Conference on Computer Vision (ECCV)}, 2018, pp. 568--583.

\bibitem{zhao2021graphsmote}
T.~Zhao, X.~Zhang, and S.~Wang, ``Graphsmote: Imbalanced node classification on
  graphs with graph neural networks,'' in \emph{Proceedings of the 14th ACM
  international conference on web search and data mining}, 2021, pp. 833--841.

\bibitem{chen2021topology}
D.~Chen, Y.~Lin, G.~Zhao, X.~Ren, P.~Li, J.~Zhou, and X.~Sun,
  ``Topology-imbalance learning for semi-supervised node classification,''
  \emph{Advances in Neural Information Processing Systems}, vol.~34, pp.
  29\,885--29\,897, 2021.

\bibitem{wang2022ada}
Y.~Wang, Y.~Zhang, F.~Zhang, S.~Wang, M.~Lin, Y.~Zhang, and X.~Sun, ``Ada-nets:
  Face clustering via adaptive neighbour discovery in the structure space,''
  \emph{arXiv preprint arXiv:2202.03800}, 2022.
  
\bibitem{he2009learning}
H.~He and E.~A. Garcia, ``Learning from imbalanced data,'' \emph{IEEE
  Transactions on knowledge and data engineering}, vol.~21, no.~9, pp.
  1263--1284, 2009.

\end{thebibliography}
\end{document}